\relax
\documentclass[12pt]{article}

\usepackage{times}
\usepackage{helvet}
\usepackage{courier}
\usepackage[usenames,dvipsnames]{color}
\usepackage[table,x11names]{xcolor}
\definecolor{lightgray}{gray}{0.90}
\usepackage{tikz}

\newcommand\copyrighttext{%
  \footnotesize Published as a 2-page research abstract at the International Symposium on Combinatorial Search (SoCS), 2016}
\newcommand\copyrightnotice{%
\begin{tikzpicture}[remember picture,overlay]
\node[anchor=south,yshift=-40pt] at (current page.south) {\fbox{\parbox{\dimexpr\textwidth-\fboxsep-\fboxrule\relax}{\copyrighttext}}};
\end{tikzpicture}%
}

\usepackage{amsmath}
\usepackage{amsfonts}
\usepackage{enumitem}

\usepackage{wrapfig}
\usepackage{scrextend}
\usepackage{graphicx}
\usepackage{breakcites}

\usepackage{multirow}
\usepackage{hhline}
\usepackage{amssymb}   
\usepackage{fancyvrb}   

\usepackage{algorithm}

\usepackage{geometry}
\usepackage[noend]{algpseudocode}
\algrenewcommand\alglinenumber[1]{\tiny #1:}

\usepackage{caption}
\makeatletter
\def\BState{\State\hskip-\ALG@thistlm}
\makeatother

\usepackage{hyperref}
\hypersetup{
     colorlinks   = true,
     citecolor    = blue,
     linkcolor    = blue,
     urlcolor     = blue
}

\newcommand \cc {\texttt{\textsc{OP-COUNT}}}
\newcommand{\oppa}{\texttt{OP-COUNT} }

\begin{document}
%
\title{Compliant Conditions for Polynomial Time Approximation of Operator Counts}
\author{
Tathagata Chakraborti$^*$ ~~~~ Sarath Sreedharan$^*$ ~~~ Sailik Sengupta$^*$
\\ T. K. Satish Kumar$^\dagger$ ~~~~ Subbarao Kambhampati$^*$\\
{\small}\\
$^*$Dept. of Computer Science, Arizona State University\\
$^\dagger$Dept. of Computer Science, University of Southern California\\
$^*${\tt \small{\{tchakra2, ssreedh3, sailiks, rao\}@asu.edu }}  $^\dagger${\tt \small tkskwork@gmail.com}\\
\date{}
}

\maketitle
\copyrightnotice
\begin{abstract}
In this paper, we develop a computationally simpler version of the operator count heuristic for a particular class of domains. The contribution of this abstract is threefold, we (1) propose an efficient closed form approximation to the operator count heuristic using the Lagrangian dual; (2) leverage compressed sensing techniques to obtain an integer approximation for operator counts in polynomial time; and (3) discuss the relationship of the proposed formulation to existing heuristics and investigate properties of domains where such approaches appear to be useful.
\end{abstract}

\subsection{The \oppa Heuristic}

%

\subsubsection{Domain Model.} The domain is described by a set of variables $f\in\mathcal{F}$ which can assume values from a (finite) domain $D(f) \subseteq \mathbb{N}$. A state is given by the particular assignment of values to these variables: $\mathbb{S} = \{f = v~|~ v \in D(f)~\forall f \in \mathcal{F}\}$. The value of variable $f$ in state $\mathbb{S}$ is referred to as $\mathbb{S}(f)$. 
The action model $\mathcal{A}$ consists of operators $a = \langle C_a, E_a\rangle$ where $C_a$ is the cost of the action, 
and $E_a = \{\langle f, v_o, v_n \rangle~|~f\in\mathcal{F}; v_o, v_n\in \{-1\}\cup D(f)\}$ is the set of effects. The transition function $\delta(\cdot)$ determines the next state after the application of action $a$ to state  $\mathbb{S}$ as - 
{\small
\begin{align*}
\delta(a, \mathbb{S}) = \bot \text{ if } \exists \langle f, v_o, v_n\rangle \in E_a \text{ s.t. } v_o \not= -1 \wedge v_o\not= \mathbb{S}(f);\\
= \{f = v_n \forall \langle f, v_o, v_n\rangle \in E_a; \text{ else } f = \mathbb{S}(f)\} \text{ otherwise.}
\end{align*}
}%
\subsubsection{Plans and Operator Counts.} 
A planning problem is a tuple $\Pi = \langle \mathcal{F}, \mathcal{A}, \mathbb{I}, \mathbb{G} \rangle$, where $\mathbb{I}, \mathbb{G}$ are the initial and (partial) goal states respectively.
The solution to the planning problem is a \emph{plan} $\pi = \langle a_1, a_2, \ldots \rangle,~\pi(i)=a_i \in \mathcal{A}$ such that $\delta(\pi, \mathbb{I}) \models \mathbb{G}$, where the cumulative transition function is given by $\delta(\pi, \mathbb{S}) = \delta(\langle a_2, a_3, \ldots\rangle, \delta(a_1, \mathbb{S}))$. The cost of the plan is given by $C(\pi) = \sum_{a \in \pi}C_a$ and an \emph{optimal plan} $\pi^*$ is such that $C(\pi^*) \leq C(\pi)~\forall\pi$.
The operator count for an action $a$ given a plan $\pi$ is given by $\lambda(a,\pi) = |\{i~|~a = \pi(i)\}|$ and the total operator count of the plan $\lambda(\pi) = |\pi|$. 

\subsubsection{Compliant Variables.} We define compliant variables as those that whenever they occur as a precondition of an action, they must also be an effect, and vice versa. Thus, $f \in \mathcal{F}$ is \emph{compliant} iff $\forall a \in \mathcal{A}, \langle f, v_o, v_n\rangle \in E_a \implies v_o \not=-1 \wedge v_n \not=-1$; $f$ is referred to as \emph{rogue} otherwise. Let $\Phi \subseteq \mathcal{F}$ be the set of all compliant variables, and the set of compliant variables whose values are specified in the goal be $\phi \subseteq \Phi$, henceforth referred to as goal compliant conditions.

\subsubsection{The State Transformation Equation.}
Let $|\phi| = m$ and $|\mathcal{A}| = n$.
Consider an $m\times n$ matrix $\mathbf{M}$ whose $ij^{th}$ element $M_{ij} \in \mathbb{Z}$ is the numerical change in $f_i \in \phi$ produced by action $a_j \in \mathcal{A}$, i.e. $M_{ij} = v_n - v_o;~\langle f_i, v_o, v_n \rangle \in E_{a_j}$.
Also, let $\mathbf{D}$ be a vector of size $m$ whose $i^{th}$ entry $d_i$ is the change in a goal compliant $f \in \phi$ from the current state to the final state, i.e. $d_i = v_g - v_c; v_g = f_i \in \mathbb{G}, v_c = f_i \in \mathbb{S}$; and let $\mathbf{x}$ be a vector of size n, whose $i^{th}$ element is $x_i \in \mathbb{N}$.
Then the following equality holds:
\begin{align}
\label{11}
\mathbf{M}\mathbf{x} & = \mathbf{D} 
\end{align}
The integer solution $\mathbf{x}^*$ to this system of linear equations with the least $|\mathbf{x}^*|$ gives a lower bound on the operator counts required to solve the planning problem, i.e. $|\mathbf{x}^*| \leq |\pi^*|$.  We can compute a real-valued approximation in closed-form, by
\begin{align}
    \min~~~&||\mathbf{Q}\mathbf{x}||_2^2\\
    s.t.~~~\mathbf{M}\mathbf{x} &= \mathbf{D}
\end{align}
using the Lagrangian multiplier method for this optimization problem as follows -
{\begin{align}
\label{101}
L(\mathbf{x}) &= \frac{1}{2}||\mathbf{Q}\mathbf{x}||^2 + \lambda^T(\mathbf{D} - \mathbf{M}\mathbf{x})\\
\implies \mathbf{x}^* &= \mathbf{Q}^{-2}\mathbf{M}^T(\mathbf{M}\mathbf{Q}^{-2}\mathbf{M}^T)^{-1}\mathbf{D} 
\end{align}}
Here $\mathbf{Q}$ is a $n\times n$ matrix of action costs whose $ij^{th}$ entry $Q_{ij} = C_{a_i} \text{ if } i = j;~0 \text{ otherwise}$ (for unit cost domains) $\mathbf{Q}$ is an identity matrix and
$\mathbf{x}^* = \mathbf{M}^T(\mathbf{M}\mathbf{M}^T)^{-1}\mathbf{D}$
The most costly operation here is the calculation of the pseudo inverse, which can be done in $\approx \mathcal{O}(n^{2.3})$ time. Further, $\mathbf{M}$ is problem independent, and hence the factor 
$\mathbf{Z} = \mathbf{Q}^{-2}\mathbf{M}^T(\mathbf{M}\mathbf{Q}^{-2}\mathbf{M}^T)^{-1}$ can be \textit{precomputed} given an action model.
Thus it follows that we can readily use $||\mathbf{QZD}||$ as a heuristic for state-space search.

Note that this formulation can also determine infeasibility of goal reachability immediately (in domains where actions are not reversible this is extremely useful) when the system is unsolvable, as shown in Algorithm \ref{algo}.  Unfortunately, the use of the $l_2$-\emph{norm}, that helps us in obtaining the closed-form polynomial bound heuristic, also makes the heuristic inadmissible.

\begin{algorithm}[btp]

\caption{Using \oppa Heuristic for State-Space Search}
\label{CompCount}
\label{algo}
\begin{algorithmic}
\Procedure{Pre-compute}{$\Pi$}
\State Compute $\mathbf{M}, \mathbf{Q}$
\State Convert $\mathbf{M}$ to row echelon form $\rightarrow \mathbf{T}$ is the transformation matrix, $r$ is the rank
\State $\mathbf{Y} \leftarrow \mathbf{M}[1:r, :],~\mathbf{Z} \leftarrow \mathbf{Q}^{-2}\mathbf{Y}^T(\mathbf{Y}\mathbf{Q}^{-2}\mathbf{Y}^T)^{-1}$
\EndProcedure
\vspace{3pt} 
\Procedure{$h(\mathbb{S}) = $ \cc}{$\mathbb{S}, \mathbb{G}$}
\State Compute $\mathbf{D}=\mathbf{G}-\mathbf{S}$
\State Compute $T^d = \mathbf{T} \times \mathbf{D}$ and $\tau = \mathbf{T}^d[1:r]$
\If {$t^d_i \not= 0~\forall i \geq r+1$} \textit{No solution!} 
\Else \indent return $\lceil\mathbf{Q\times\mathbf{Z}\times\tau}\rceil$
\EndIf
\EndProcedure
\end{algorithmic}
\end{algorithm}

\subsubsection{Sparse coding.} 
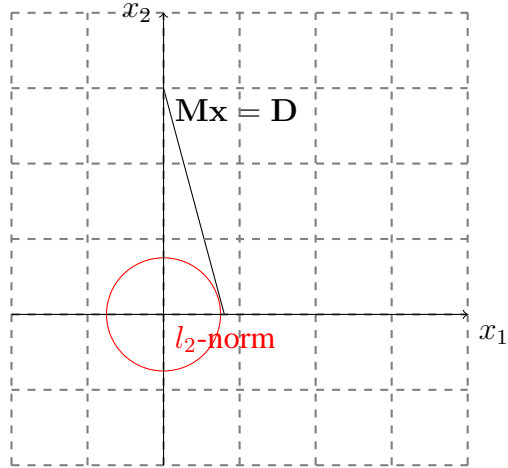
\begin{figure}[!h]
\centering
    \begin{tikzpicture}
    \draw (0,3) node[below right] {$\mathbf{Mx=D}$} -- (0.8,0);
    \draw (0,0)[red] circle (0.75cm) node[below right]{$l_2$-norm};
    \draw[thick,color=gray,step=1cm,
    dashed] (-2,-2) grid (4,4);
    \draw[->] (-2,0) -- (4,0)
    node[below right] {$x_1$};
    \draw[->] (0,-2) -- (0,4)
    node[left] {$x_2$};
    \end{tikzpicture}
    \caption{Eucledian norm minimization produces small fractional values for $x_1$ and $x_2$}
    \label{fig:1}
\end{figure}

Since operator counts are integers, we would ideally want an integer solution to Eqn \ref{101} (which makes the problem computationally intractable).  Unfortunately, the polynomial bound Lagrangian method described above does not address this aspect giving rise to bad heuristic values for certain section of problems. To describe this problem geometrically, we consider a planning domain with two compliant operators (of unit cost), such that $\mathbf{x}=<x_1, x_2>$.  If the plane inscribed by $\mathbf{Mx=D}$ in the two dimensional space is close two either of the axis, the $l_2$ norm calculated above results in small fractional values, and hence a less informed heuristic. As can be seen in the figure \ref{fig:1}, the actual operator counts for the given example (with $M=\begin{pmatrix} 15 & 4 \end{pmatrix}$ and $D= \begin{pmatrix} 12 \end{pmatrix}$) should have been $x_1=0$ and $x_2=3$.  But the $l_2$ minimization results in small fractional values with $x_1=0.77$ and $x_2=0.77$, and the heuristic values of $h_{l_2}=1.54$ instead of $|\pi^*|=3$.

Thus, we propose a different approximation method to obtain integer values for individual operator counts, remaining within the polynomial time bound.

We notice that in most cases $n\gg m$ and also $n \gg |\mathbf{x}^*|$ due to the combinatorial explosion during grounding of domains.  Thus, we propose an operator count heuristic that exploits this knowledge about the sparsity of $\mathbf{x}^*$.  Ideally, we would like to solve the following problem,

\begin{eqnarray}
\min && |\mathbf{x}|_{l_0} \nonumber \\
s.t.~~~~~Mx &=& D \nonumber \\
\mathbf{x} &\succeq& 0 \nonumber
\end{eqnarray}

\noindent since minimizing the $l_0$ norm results in the sparsest solution.  But, we encounter two problems.  Firstly, the optimal operator counts ($\mathbf{x}^*$), although sparse, might not be the sparsest solution.  Secondly, minimizing the $l_0$ norm is \emph{NP}-hard~\cite{ge2011note}.

Thus, we draw upon compressed sensing techniques to enforce a level of sparsity when computing the vector $\mathbf{x}$.  To this end, we suggest minimization of \textit{$l_1$-norm} (\textit{$l_1$-LP}) or weighted \textit{$l_1$-norm} (\textit{$\omega$-$l_1$-LP}) \cite{candes2008enhancing} to enforce positive integer solutions.

\begin{figure}[!h]
\centering
    \begin{tikzpicture}
    \draw (0,3) node[below right] {$\mathbf{Mx=D}$} -- (0.8,0);
    \draw[red, rotate around={45:(0,0)}] (-0.56,-0.56) rectangle (0.56,0.56) node[below right] {$l_1$ norm};
    \draw[thick,color=gray,step=1cm,
    dashed] (-3,-3) grid (4,4);
    \draw[->] (-2,0) -- (4,0)
    node[below right] {$x_1$};
    \draw[->] (0,-2) -- (0,4)
    node[left] {$x_2$};
    \end{tikzpicture}
    \begin{tikzpicture}
    \draw (0,3) node[below right] {$\mathbf{Mx=D}$} -- (0.8,0);
    \draw[red] (0,3) -- (0.4,0) node[below right] {$\omega$-$l_1$ norm};
    \draw[red] (0,3) -- (-0.4,0);
    \draw[red] (0,-3) -- (0.4,0);
    \draw[red] (0,-3) -- (-0.4,0);
    \draw[thick,color=gray,step=1cm,
    dashed] (-3,-3) grid (4,4);
    \draw[->] (-2,0) -- (4,0)
    node[below right] {$x_1$};
    \draw[->] (0,-2) -- (0,4)
    node[left] {$x_2$};
    \end{tikzpicture}
    \caption{Eucledian norm minimization produces small fractional values for $x_1$ and $x_2$}
    \label{fig:2}
\end{figure}
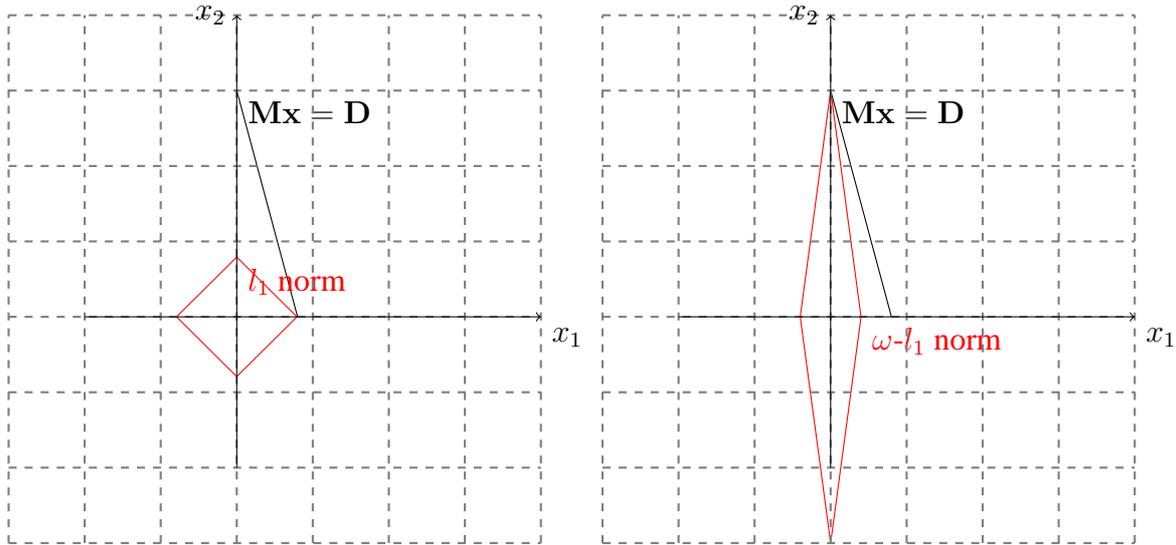

Geometrically, as can be seen in figure \ref{fig:2} these norms produce a more informed heuristic ($h_{l_1}=1.60$ and $h_{\omega-l_1}=3.4$) for the aforementioned problem.  This method tries to compress (minimize) the norm ball (or box for that matter) as much as possible till it fits in the plane $\mathbf{Mx=D}$.  The operator (dimension) that induces a tighter constraint ($x_1$ in our case), limits the expansion of the norm ball, producing a less informed heuristic ($h_{l_1}=1.60$).  The weighted \textit{$l_1$-norm} method addresses this problem by minimizing the \textit{$l_1$-norm} and iteratively penalizing the increase along the tightest dimension till convergence is reached or maximum number of iterations are achieved, resulting in a more informed heuristic ($h_{\omega-l_1}=3.4$).

For \textit{$\omega$-$l_1$-LP}, we empirically observe that rounding up the individual operator counts produce a more informed heuristic.  Thus, we arrive at a polynomial time proxy for integer solutions.

\subsubsection{Evaluations.}
The table shows the evaluation of the proposed heuristics across a total of 83 problems from five well-known unit cost planning domains.  Each entry in the table represents the percentage difference in the initial state heuristic value and the optimal plan length averaged across the problems in each domain.  The \%-compliance column shows the average number of goal compliant predicates in the problems.  Rows 1-3 show the performance of our heuristic on the original domains (`-' indicates that the heuristics could not be computed due to absence of any goal complaint variables).  Rows 3-6 show the performance in domains where the $\%$-compliance was increased (this was done by identifying instances in the action model where variables assume a don't care condition, i.e. a value of -1, and replacing it with appropriate values as entailed by domain axioms). Finally, rows 6-9 show the performance of our heuristics in problems with more completely specified goals (which results in higher percentage compliance).  As expected, our heuristic performs better as $\%$-compliance increases across a particular domain.  The performance of $l_1$ LP and $\omega$-$l_1$ LP highlights the usefulness of compressed sensing techniques in obtaining better integer approximations to the MILP.

\begin{table}[h!]
\centering
\label{eval}
\begin{tabular}{ c || c  || c  | c | c | c }
  \hline    
  Domains & \%-compliance & $l_1$-MILP & $l_1$-LP & $\omega-l_1$-LP & \cc \\
  \hline            
  \hline            
   GED & 34.29$\%$ & \cellcolor{yellow!30} 55.48$\%$    & \cellcolor{yellow!30} 55.48$\%$    & 75.76$\%$ & 55.48$\%$ \\
  \hline
   Blocks-3ops & 31.25$\%$ &    47.80$\%$ &    47.80$\%$ &    \cellcolor{yellow!30} 23.60$\%$ &    52.60$\%$ \\
  \hline            
   Blocks-4ops & 19.64$\%$ &    67.71$\%$ &    67.71$\%$ &    \cellcolor{yellow!30} 35.42$\%$ &    67.71$\%$ \\
  \hline            
   Visitall& -& -& -& -& -\\
  \hline
  \hline
   \rowcolor{lightgray} GED & 25.49$\%$ & 37.61$\%$ &  \cellcolor{yellow!15}  34.02$\%$ &    53.36$\%$ &    48.32$\%$ \\            
  \hline
   \rowcolor{lightgray} Blocks-3ops & 31.25$\%$ &    47.80$\%$ &    47.80$\%$ &  \cellcolor{yellow!15}  23.60$\%$ &    52.60$\%$ \\
  \hline            
   \rowcolor{lightgray} Blocks-4ops & 19.64$\%$ &    67.71$\%$ &    67.71$\%$ &  \cellcolor{yellow!15}  35.42$\%$ &    67.71$\%$ \\
  \hline            
   \rowcolor{lightgray} Visitall & 21.75$\%$ & \cellcolor{yellow!15} 28.41$\%$ &    \cellcolor{yellow!15} 28.41$\%$ &    44.37$\%$ &    100.00$\%$ \\
  \hline
  \hline            
   Blocks-3ops & 48.13$\%$ &    \cellcolor{yellow!30} 28.68$\%$ &    \cellcolor{yellow!30} 28.68$\%$ &    44.38$\%$ &    32.32$\%$ \\
  \hline            
   Blocks-4ops & 42.86$\%$ &    56.25$\%$ &    56.25$\%$ &    \cellcolor{yellow!30} 12.50$\%$ &    64.58$\%$ \\
  \hline            
    8-puzzle & 88.89$\%$ & \cellcolor{yellow!30} 33.33$\%$ & 40.00$\%$ &    46.67$\%$ &    40.00$\%$  \\
  \hline            
  \hline  
\end{tabular}
\end{table}

\subsection{Discussion and Related Work}

\subsubsection{Relation to Existing Heuristics.} The proposed heuristic has close associations with both heuristics on state change equations and operator counts~\cite{pommerening2014lp, bonet2014flow, van2007lp}. 
Specifically, compliant conditions capture the net change criteria very succinctly and are thus extremely useful where such properties are relevant. 
Another interesting connection to existing work is with respect to graph-plan based heuristics \cite{blum1997fast}, except here we are relaxing preconditions instead of delete effects.

\subsubsection{Compliance.} Our approach works better in domains that have many goal compliant conditions, e.g. in manufacturing domains \cite{nau1995ai} or in puzzles like Sudoku \cite{babu2010linear}. Thus goal completion strategies and semantic preserving actions have a direct effect on the quality of the heuristic. 
Intermediate representations such as transition normal form (TNF) \cite{pommerening2015normal} should be investigated in this context.

\subsubsection{Landmarks.} Our purpose here is not to compete with the most sophisticated heuristics of today but to motivate a special case that can be computed extremely efficiently. We discussed the simplest version of this formulation here, but it can be easily extended to incorporate more informative features like \emph{landmarks} \cite{porteous}. A landmark constraint is added by simply subtracting the corresponding net change from $\mathbb{D}$:
$d_i \leftarrow d_i - k_a\times(x_n-x_o)$ if $\langle d_i, x_o, x_n\rangle \in E_{a} \text{ and } a \in \mathcal{A}\text{ is an action landmark}$ with cardinality $k_a$;
and the closed form solution remains valid. In fact in terms of plan recognition with operator counts, observations are landmarks and the same approach applies. This demonstrates the flexibility of our approach.

\subsubsection{Resource Constrained Interaction.} The approach is especially relevant in the context of multi-agent interactions constrained by usage $\pi^\alpha(\eta)$ of a shared resource $\eta$ by a plan $\pi^\alpha$ of an agent $\alpha$. For example, in an adversarial setting, if an agent $\alpha_2$  wanted to stop $\alpha_1$ from executing its plan, all it needs to do is to ensure that $\exists\eta \text{ s.t. } \pi^{\alpha_1}(\eta) + \pi^{\alpha_2}(\eta) > |\eta|$. Similarly, in a cooperative setting, if agent $\alpha_2$ wanted to ensure that $\alpha_1$'s plan succeeds, it would need to make sure that $\forall\eta~\pi^{\alpha_1}(\eta) + \pi^{\alpha_2}(\eta) \leq |\eta|$. In fact, as resource variables are compliant, our approach may provide quick estimates of an agent's intent without computing the entire plan.

\vspace{5pt}
{\small
\noindent\emph{Acknowledgment.} This research is supported in part by the ONR grants N00014-13-1-0176, N00014-13-1-0519 and N00014-15-1-2027, and ARO grant W911NF-13-1-0023.
}

\bibliographystyle{plain}
\bibliography{single_col}

\end{document}